\documentclass[times]{TRR}

\usepackage{moreverb,url}

\usepackage[colorlinks,bookmarksopen,bookmarksnumbered,citecolor=red,urlcolor=red]{hyperref}

\newcommand\BibTeX{{\rmfamily B\kern-.05em \textsc{i\kern-.025em b}\kern-.08em
T\kern-.1667em\lower.7ex\hbox{E}\kern-.125emX}}

\def\volumeyear{2020}

\begin{document}

\runninghead{Smith and Haynes}

\title{Control and State Estimation of Vehicle-Mounted Aerial Systems in GPS-Denied, Non-Inertial Environments}

\author{Riming Xu\affilnum{1}, Obadah Wali\affilnum{1}, Yasmine Marani\affilnum{1}, Eric Feron\affilnum{1}}

\affiliation{\affilnum{1}the Robotics, Intelligent Systems, and Control Lab, King Abdullah University of Science and Technology (KAUST), Thuwal, Saudi Arabia.}

\corrauth{Riming Xu, riming.xu@kaust.edu.sa}

\begin{abstract}
We present a robust control and estimation framework for quadrotors operating in Global Navigation Satellite System(GNSS)-denied, non-inertial environments where inertial sensors such as  Inertial Measurement Units (IMUs) become unreliable due to platform-induced accelerations. In such settings, conventional estimators fail to distinguish whether the measured accelerations arise from the quadrotor itself or from the non-inertial platform, leading to drift and control degradation. Unlike conventional approaches that depend heavily on IMU and GNSS, our method relies exclusively on external position measurements combined with a Extended Kalman Filter with Unknown Inputs (EKF-UI) to account for platform motion. The estimator is paired with a cascaded PID controller for full 3D tracking. To isolate estimator performance from localization errors, all tests are conducted using high-precision motion capture systems. Experimental results in a moving-cart testbed validate our approach under both translational in X-axis and Y-axis dissonance. Compared to standard EKF, the proposed method significantly improves stability and trajectory tracking without requiring inertial feedback, enabling practical deployment on moving platforms such as trucks or elevators.
\end{abstract}

\maketitle

\section{Introduction}
{U}{nmanned} Aerial Vehicles (UAVs) have been extensively deployed in a variety of applications such as aerial imaging, inspection, search and rescue, and logistics~\cite{gao2024aerial}~\cite{mohsan2022towards}~\cite{saadiyean2024learning}~\cite{qi2024minimizing}~\cite{wali2023non}. 
However, most existing control strategies assume that UAVs operate in an inertial frame and rely on GNSS- or IMU-based navigation systems, whether using conventional model-based control~\cite{falanga2019fast}~\cite{wei2025meta}~\cite{ren2025safety} or emerging learning-based approaches~\cite{song2023reaching}~\cite{nobrega2024proximal}~\cite{hwangbo2017control}.
As UAV deployment expands to more diverse scenarios, it is increasingly likely that Unmanned Aerial Vehicles (UAVs) will operate inside accelerating platforms such as trucks, ships, or elevators, environments characterized by non-inertial motion. In such cases, GNSS signals are typically denied.
These emerging use-cases increasingly demand UAVs to function reliably in GPS-denied, non-inertial environments that were previously considered out of scope for aerial autonomy.

Moreover, conventional control frameworks based on Kalman filtering can become highly unreliable in these settings, since IMU measurements conflate UAV motion with non-inertial platform dynamics, while the filter itself assumes disturbances are random noise rather than structured accelerations. This mismatch results in biased estimates and drift. In addition, the dynamic accelerations imparted by the non-inertial platform can intensify the propagation of inherent bias and drift in IMU readings, especially in scenarios where the moving platform imparts rapidly varying accelerations that are not directly measurable by the UAV, leading to ambiguity in state estimation. This is because the onboard IMU may incorrectly assume the UAV is stationary when, in fact, its relative position is changing due to the motion of the none-inertial platform. Such sensor inconsistency introduces significant estimation errors and leads to prolonged convergence times for the filter, compromising both stability and control accuracy~\cite{marani2022drone}. More critically, standard filtering frameworks such as the Extended Kalman Filter are fundamentally ill-suited for such settings. They typically model disturbances as zero-mean Gaussian noise, and cannot capture the low-frequency, structured accelerations caused by platform motion resulting in biased estimates.

So far, few works have explored how UAVs behave in non-inertial environments. Examples include the work in ~\cite{lu2018shipboard}, which developed a relative-coordinate controller for autonomous shipboard landing under unknown platform motion and external disturbances, without requiring inertial frame measurements. Tognon et al. propose in ~\cite{tognon2016observer} an observer-based nonlinear controller that enables exact position and tension tracking of a tethered UAV relative to a generically moving platform without assuming platform dynamics are known or controllable. A related work in ~\cite{alami2020design} investigates the tracking challenges of a UAV operating in a non-inertial reference frame, utilizing a cascaded PID control strategy. 
Recent works have taken significant steps toward enabling UAV operation in more dynamic and unpredictable environments, offering insights that support this study. Curvilinear trajectory tracking under high-speed pursuit is investigated in~\cite{gupta2025curvitrack}, which emphasizes precise estimation under fast non-linear target motion. Reinforcement learning has been applied to swarm obstacle avoidance under unknown environments~\cite{zhang2024cooperative}, and cooperative fencing control using only relative sensing was developed in~\cite{zhao2024cooperative}. In the realm of dynamic planning, Xu et al. proposed intent-driven model predictive control~\cite{xu2025intent} and reinforcement learning-based safe navigation~\cite{xu2025navrl} frameworks for real-time adaptation.

Visual self-localization using DINOv2-based deep features enables robust performance in low-altitude urban scenes~\cite{yang2025dinov2}, while perception and planning pipelines that can rapidly adapt in cluttered scenarios are addressed in~\cite{lu2024fapp}. Ghotavadekar et al.~\cite{ghotavadekar2024variable} present variable time-step MPC tailored for interception of moving targets, and swarm-based source-seeking using onboard control is demonstrated in~\cite{karaguzel2023shadows}. Foundational work by Herissé et al.~\cite{herisse2011landing} laid early groundwork for optical flow-based landing on moving platforms.
To the best of our knowledge, there is limited work in the literature on UAVs operating entirely inside moving environments without relying on inertial localization. In a non-inertial frame, if a classical filter treats the carrier’s low-frequency, structured accelerations as zero-mean noise, it will misattribute them to the vehicle’s motion, producing systematic bias and drift, leading to ambiguous measurements. Conventional estimators assume that vehicle accelerations are separable from external disturbances and can be modeled as stochastic noise. In non-inertial frames, however, platform-induced accelerations are structured and persistent, causing systematic bias. In this paper, we formulate the control problem in a non-inertial reference frame and present an improved state estimator that accounts for non-inertial dynamics, together with a minimalistic controller. The UAV is controlled using relative localization within the moving frame, which typically relies on vision-based sensing. However, for the purpose of validating our proposed framework and isolating control performance from localization errors, we employ two high-precision motion capture systems (MoCap) as the localization source, along with supporting simulations and experimental validations. This deliberate choice enables us to decouple the effects of localization from those of control, ensuring that observed performance differences are attributable to control dynamics rather than perception noise. While vision-based localization is ultimately needed for deployment, isolating and validating the controller under idealized conditions is a necessary precursor to more complete autonomy.

While UAV dynamics have traditionally been modeled using the Newton–Euler formalism in inertial frames~\cite{bouabdallah2007design}~\cite{hoffmann2007quadrotor}~\cite{luo2024novel}, more recent approaches have explored structure-preserving and geometry-aware formulations, such as port-Hamiltonian frameworks on Lie groups~\cite{duong2024port}, which offer improved physical consistency and generalization to complex robotic systems. However, both types of models typically assume an inertial reference frame, and may become insufficient when the UAV operates inside a moving environment.
Traditional formulations assume that the inertial acceleration experienced by the UAV is solely caused by its own actuation, whereas in a non-inertial frame, the motion of the environment introduces additional apparent forces that must be accounted for explicitly. In such cases, direct application of standard models can result in significant estimation and control inaccuracies.

These studies demonstrate the increasing need to integrate robustness to environmental dynamics, particularly when the UAV platform is denied inertial reference. However, many such frameworks still operate under inertial assumptions or rely heavily on onboard IMU/GNSS data. This motivates our study on robust UAV control inside non-inertial, GPS-denied environments using external-only sensing.

However, existing works remain limited in several ways. Many still rely on IMU or GNSS inputs, which become unreliable under non-inertial accelerations. Others address platform motion only under restrictive assumptions, such as known or controllable dynamics, or require tethering. Moreover, conventional approaches often separate estimation and control, allowing biased state estimates to directly degrade stability.

In contrast, our contribution is twofold: (i) we propose an IMU/GNSS-free control framework that relies solely on external position sensing, augmented by an EKF with Unknown Inputs (EKF-UI) to explicitly model and compensate platform-induced accelerations; and (ii) we integrate this estimator tightly with a cascaded PID controller, enabling robust 3D tracking inside GPS-denied, non-inertial environments. To our knowledge, this is the first work to experimentally validate such a framework in a moving-cart testbed with both translational and rotational dissonance.

The remainder of this paper is organized as follows: Section II introduces the estimation dynamics, Section III presents simulation results, Section IV describes the controller adjustment, Section V details the experimental setup, Section VI discusses the results, and Section VII concludes the paper.

\section{the estimation dynamics}

There are several limitations of the conventional Extended Kalman Filter (EKF) in non-inertial reference frames, such as the misinterpretation of external accelerations, breakdown of the static environment assumption, and lack of platform motion awareness. In non-inertial environments, the IMU records absolute accelerations that conflate the UAV’s self-induced motion with the platform’s acceleration, making relative motion unobservable without additional sensing. Visual sensors assume a static environment, which fails when the robot is on a moving platform. The EKF then incorrectly merges visual and inertial data, degrading the accuracy of state estimation. EKF does not include the motion of the platform in its state representation. Without modeling the platform dynamics, the EKF misinterprets these external forces as self-motion, leading to estimation drift. This limits its adaptability in dynamic environments, where reference frames are themselves in motion.

We begin by deriving the equations of motion of the drone in an inertial frame, and subsequently apply relative motion principles to obtain its dynamics in a non-inertial reference frame. This enables explicit modeling of the platform's motion, thereby allowing the estimator to distinguish between vehicle-induced accelerations and those arising from external platform dynamics, which are otherwise conflated in conventional formulations.

The equations of motion of the drone in the inertial frame are given by Equation~\eqref{eq:u_vect}:

\begin{equation}
\label{eq:u_vect}
\begin{aligned}
\begin{bmatrix}
\ddot{x} \\
\ddot{y} \\
\ddot{z} \\
\ddot{\phi} \\
\ddot{\theta} \\
\ddot{\psi}
\end{bmatrix} =\ & 
\underbrace{
\begin{bmatrix}
\cos\phi \sin\theta \cos\psi + \sin\phi \sin\psi \\
\cos\phi \sin\theta \sin\psi - \sin\phi \cos\psi \\
\cos\phi \cos\theta
\end{bmatrix} u_1
+
\begin{bmatrix}
0 \\ 0 \\ -g
\end{bmatrix}
}_{\text{Position acceleration part}}
\\[10pt]
&\oplus\ 
\underbrace{
\left(
\mathbf{M}
\begin{bmatrix}
\theta \dot{\psi} \\
\phi \dot{\psi} \\
\theta \dot{\phi}
\end{bmatrix}
+
\begin{bmatrix}
u_2 \\ u_3 \\ u_4
\end{bmatrix}
\right)
}_{\text{Angular acceleration change part}}
\end{aligned}
\end{equation}

where 
$\ddot{x}$, $\ddot{y}$, $\ddot{z}$ is the translational acceleration of the UAV along the inertial $x$-axis, $y$-axis, $z$-axis.
and the coupling matrix $\mathbf{M}$, which accounts for the gyroscopic effects due to the differences in principal moments of inertia, is given by Equation~\eqref{eq:inertia_matrix}:

\begin{equation}
\label{eq:inertia_matrix}
\mathbf{M} =
\begin{bmatrix}
\frac{I_y - I_z}{I_x} & 0 & 0 \\
0 & \frac{I_z - I_x}{I_y} & 0 \\
0 & 0 & \frac{I_x - I_y}{I_z}
\end{bmatrix}.
\end{equation}

Here, $I_x$, $I_y$, and $I_z$ denote the moments of inertia about the body-fixed $x$-, $y$-, and $z$-axes, respectively.

In these formulations, $g$ denotes the gravitational acceleration, and $\phi$, $\theta$, and $\psi$ represent the roll, pitch, and yaw angles, respectively. The $\oplus$ symbol indicates the concatenation of two vectors. The structure clearly separates the contributions of thrust (affecting translational motion) and rotational moments (affecting attitude dynamics), providing a modular and interpretable representation of the quadrotor system.

where \( u_1, u_2, u_3, u_4 \) are the inputs of the drone given by:

\begin{equation}
\label{eq:u_combined}
\mathbf{u}
= 
\begin{bmatrix}
u_1 \\
u_2 \\
u_3 \\
u_4
\end{bmatrix}
=
\begin{bmatrix}
\frac{b}{m} & \frac{b}{m} & \frac{b}{m} & \frac{b}{m} \\
0 & -\frac{b}{I_x} & 0 & \frac{b}{I_x} \\
-\frac{b}{I_y} & 0 & \frac{b}{I_y} & 0 \\
-\frac{l}{I_z} & \frac{l}{I_z} & -\frac{l}{I_z} & \frac{l}{I_z}
\end{bmatrix}
\begin{bmatrix}
\Omega_1^2 \\
\Omega_2^2 \\
\Omega_3^2 \\
\Omega_4^2
\end{bmatrix}
\end{equation}

where \( \Omega_i, i=1,2,3,4 \) represent the angular rates of the four rotors.

In the non-inertial frame scenario, the accelerations of the truck along the \( x \), \( y \), and \( z \)-axes are denoted as \( a_x, a_y, a_z \). Assume that it has only translation, the acceleration of the drone relative to the non-inertial frame is:

\begin{equation}
\label{eq:relative_acc_vector}
\ddot{\mathbf{r}} =
\begin{bmatrix}
\ddot{x}_r \\
\ddot{y}_r \\
\ddot{z}_r
\end{bmatrix}
=
\begin{bmatrix}
\ddot{x}_a \\
\ddot{y}_a \\
\ddot{z}_a
\end{bmatrix}
-
\begin{bmatrix}
a_x \\
a_y \\
a_z
\end{bmatrix}
\end{equation}

In Equation~\eqref{eq:relative_acc_vector}, $\ddot{\mathbf{r}} = [\ddot{x}_r, \ddot{y}_r, \ddot{z}_r]^\top$ denotes the drone’s acceleration relative to the moving platform, expressed in the inertial frame. $\ddot{\mathbf{a}} = [\ddot{x}_a, \ddot{y}_a, \ddot{z}_a]^\top$ represents the drone’s absolute acceleration in the inertial frame. $\mathbf{a}_p = [a_x, a_y, a_z]^\top$ denotes the platform's linear acceleration, also measured in the inertial frame. This formulation follows the classical relative acceleration relation in Newtonian mechanics and allows the estimator to explicitly account for external platform dynamics when interpreting drone motion.

Then the representation of the drone in the truck frame is Equation~\eqref{eq:rep}:

\begin{align}
\label{eq:rep}
\dot{\mathbf{x}} =
&\underbrace{
\begin{bmatrix}
x_2 \\
u_1 (\cos x_7 \sin x_9 \cos x_{11} + \sin x_7 \sin x_{11}) \\
x_4 \\
u_1 (\cos x_7 \sin x_9 \sin x_{11} - \sin x_7 \cos x_{11}) \\
x_6 \\
u_1 \cos x_7 \cos x_9 - g
\end{bmatrix}
}_{\mathbf{f}_1(\mathbf{x}, \mathbf{u})}
\nonumber \\
&\qquad\oplus\
\underbrace{
\begin{bmatrix}
x_8 \\
a x_{10} x_{12} + u_2 \\
x_{10} \\
b x_8 x_{12} + u_3 \\
x_{12} \\
c x_8 x_{10} + u_4
\end{bmatrix}
}_{\mathbf{f}_2(\mathbf{x}, \mathbf{u})}
+
\underbrace{
\begin{bmatrix}
0 \\ -a_x \\ 0 \\ -a_y \\ 0 \\ -a_z \\ 0 \\ 0 \\ 0 \\ 0 \\ 0 \\ 0
\end{bmatrix}
}_{\text{Non-inertial acceleration disturbance } }
\end{align}

Above System has the following global structure Equation~\eqref{eq:global}: 

\begin{equation}
\begin{aligned}
    \dot{x} &= f_c(x, u) + E a(t) \\
    y &= C x
\end{aligned}
\label{eq:global}
\end{equation}

To enable disturbance-aware estimation in non-inertial settings, we augment the system state with unknown inputs representing the platform’s accelerations. While the full quadrotor dynamics are represented by the 12-dimensional state vector $x(t)$ in (6), we define a reduced extended state vector $z(t)$ that captures only the translational motion and unknown accelerations:
\[
\zeta(t) =  \begin{bmatrix}
p_x & v_x & p_y & v_y & p_z & v_z & a_1 & a_2 & a_3
\end{bmatrix}^\top \in \mathbb{R}^9,
\]
where $p_i, v_i$ denote relative positions and velocities, and $d_i$ represent the unmeasured accelerations induced by the moving platform along each axis. This reduced formulation is sufficient for the EKF-UI implementation, which focuses on translational states while treating attitude separately.

Compared to the standard EKF, EKF-UI increases the dimensionality of the state vector but enables the filter to adapt to non-inertial disturbances through online estimation of the unknown accelerations.

The continuous-time dynamics of the simplified extended system are given in Equation~\eqref{eq:continuous}:

\begin{equation}
\label{eq:continuous}
\left\{
\begin{aligned}
    \dot{\zeta}(t) &= 
    \begin{pmatrix}
        \dot{z}(t) \\
        \dot{a}(t)
    \end{pmatrix}
    =
    \begin{pmatrix}
        f_c(z(t), u(t)) + E a(t) \\
        0
    \end{pmatrix}
    +
    \begin{pmatrix}
        w(t) \\
        w_a(t)
    \end{pmatrix}, \\
    y(t) &= H \zeta(t) + v(t)
\end{aligned}
\right.
\end{equation}

where \( a(t) = \begin{pmatrix} a_x & a_y & a_z \end{pmatrix}^T \) is the moving environment acceleration. The matrix \( E \in \mathbb{R}^{12 \times 3} \) is a sparse selection matrix with \(-1\) entries on rows 2, 4, and 6 in columns 1, 2, and 3, respectively. This represents three unknown acceleration components of the moving platform (e.g., the truck). The matrix \( H \in \mathbb{R}^{9 \times 12} \) is a sparse selection matrix with nonzero entries on rows 1, 2, 3 selecting states 1, 3, and 5, and an identity block on rows 4--9 selecting states 7--12.

The EKF with Unknown Inputs (EKF-UI) extends the traditional EKF by explicitly modeling the unknown accelerations from a moving reference frame as part of the system state. The recursive state estimation includes:

\begin{enumerate}
    \item Prediction Step: Estimate the next state using the nonlinear dynamics of the system, including control inputs and unknown inputs.
    \item Update Step: Correct the predicted state using the available position measurements to refine the estimate.
\end{enumerate}

In our onboard implementation, we use a reduced-order observer that focuses on the translational motion and unknown acceleration inputs. The extended state is defined as \( z = [p_x, v_x, p_y, v_y, p_z, v_z, d_1, d_2, d_3]^T \in \mathbb{R}^9 \), where \( p_i \), \( v_i \) are the relative position and velocity components, and \( d_i \) represent the unknown acceleration inputs along each axis.

The unknown input components $d_i$ are modeled as constant (i.e., $\dot{d}_i = 0$), and the corresponding process model Jacobian $F_k$ is computed via finite differences,

\[
F_k(:,i) \approx \frac{f(\hat{z} + \varepsilon e_i) - f(\hat{z})}{\varepsilon},
\]
where $e_i$ is the canonical unit vector in direction $i$ and $\varepsilon$ is a small constant (in our case is $10^{-6}$).

 This reduced-order model is sufficient for estimating translational velocities and external accelerations, while angular states are estimated separately using a standard EKF, as their dynamics are not affected by the non-inertial translational motion.

\subsubsection*{Prediction Step}
\begin{itemize}
    \item State Prediction:
    \[
    \hat{z}_{k|k-1} = f(\hat{z}_{k-1}, u_{k-1}) + Ts \cdot E \cdot \hat{z}_{d,k-1}
    \]
    where $f$ contains the discrete-time integrator and force model, and $T_s$ is the time step.

    \item Covariance Prediction:
    \[
    P_{k|k-1} = F_k P_{k-1} F_k^\top + Q_{\text{aug}},
    \]
    where $Q_{\text{aug}}$ includes process noise for both the core 6 states and the 3 unknown inputs.
\end{itemize}

\subsubsection*{Update Step}


The Kalman gain is computed as:
\[
K_k = P_{k|k-1} H^\top (H P_{k|k-1} H^\top + R)^{-1},
\]
and used to update the state and state covariance matrix:
\[
\hat{z}_k = \hat{z}_{k|k-1} + K_k (y_k - H \hat{z}_{k|k-1}),
\quad
P_k = (I - K_k H) P_{k|k-1}.
\]

By augmenting the state with unknown accelerations, EKF-UI enables adaptive compensation for unknown disturbances caused by the motion of the underlying platform. This is crucial in our GPS-denied, non-inertial experimental setups where platform acceleration is significant yet unmeasured.

\section{Controller adjustment}

We implement a cascaded PID controller~\cite{ren2016cascade}for stabilizing the drone in all three translational axes. The structure is divided into two loops: a horizontal controller for $x$ and $y$ axes, and a vertical controller for the $z$ axis.

The horizontal controller receives desired position commands \((x_d, y_d)\) and compares them with the current position \((x, y)\). The position errors are scaled by a proportional gain \(k_{Pxy}\) to generate desired velocities \((v_{x,\text{d}}, v_{y,\text{d}})\). These are then compared with estimated velocities \((\hat{v}_x, \hat{v}_y)\), and the resulting velocity errors are passed through a PID controller to produce desired roll and pitch commands \((u_2, u_3)\). The outputs are further rotated into the body frame based on the current yaw angle and constrained by rate limits to ensure smooth actuation.

For vertical control, a similar loop is used. The altitude error \(z_d - z\) is scaled to produce a desired vertical velocity \(v_{z,\text{des}}\), which is compared with the estimated velocity \(\hat{v}_z\). The resulting error is passed through a PID controller to compute the desired thrust. This thrust command is then converted to a PWM signal using a nonlinear interpolation function fitted from empirical motor data.

Both loops rely heavily on the quality of velocity estimates \((\hat{v}_x, \hat{v}_y, \hat{v}_z)\), which are provided by the onboard state estimator. Consequently, any bias or noise in the velocity estimation will directly affect the control performance, especially in non-inertial frames. This motivates the use of an EKF-UI estimator, as detailed in the following section.

\begin{figure}[!t]
  \centering
  \includegraphics[width=0.8\linewidth]{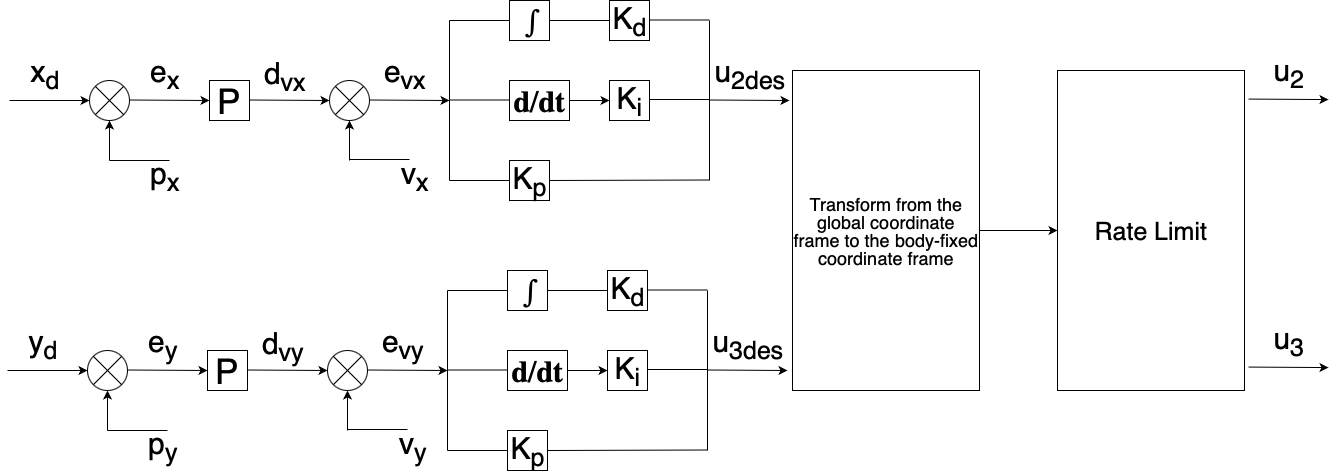}
  \caption{
    Block diagram of the horizontal PID controller for the $x$ and $y$ directions.
    Desired position is converted into velocity commands and then into body-frame attitude commands.
  }
  \label{fig:xypid}
\end{figure}

\begin{figure}[!t]
  \centering
  \includegraphics[width=0.8\linewidth]{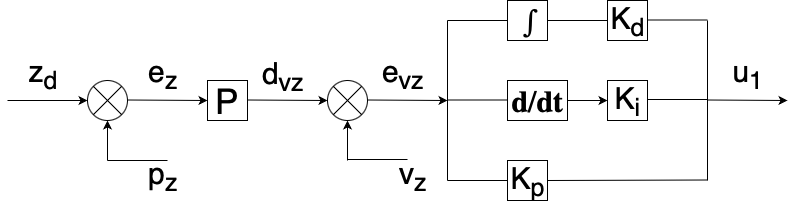}
  \caption{
    Block diagram of the vertical PID controller for the $z$ direction.
    Altitude error is processed through a cascade to compute the final thrust command.
  }
  \label{fig:zpid}
\end{figure}

Figures~\ref{fig:xypid} and \ref{fig:zpid} illustrate the control block diagrams used for horizontal and vertical regulation, respectively.

The velocity estimates provided by the state estimator directly influence the control performance in our cascade architecture. As seen in the control law implemented in our system, the horizontal velocity errors \(v_{ex} = v_{x,\text{des}} - \hat{v}_x\) and \(v_{ey} = v_{y,\text{des}} - \hat{v}_y\) are used to compute the desired roll and pitch commands (\(u_2\) and \(u_3\)) via a PID loop:
\[
u_2 \propto -\text{PID}(v_{ey}), \quad u_3 \propto \text{PID}(v_{ex})
\]

By augmenting the state with unknown acceleration terms \((d_1, d_2, d_3)\) and explicitly compensating for the platform motion, the proposed EKF-UI significantly reduces velocity estimation error in the presence of unknown disturbances. As a result, the control loop receives more accurate state feedback, leading to improved stability and reduced tracking error, especially in horizontal position regulation.



\section{Experiments and testing}

\begin{figure}[!t]
  \centering
  \includegraphics[width=0.9\linewidth]{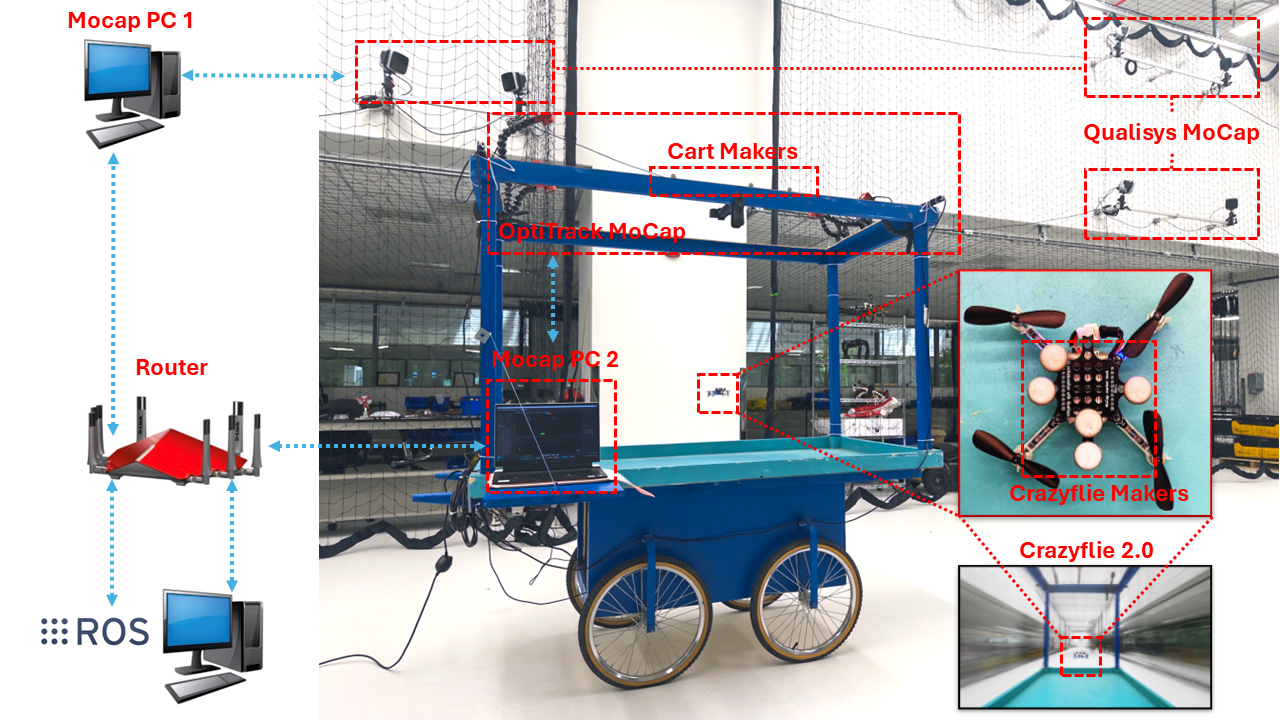}
  \caption{System overview of the proposed experimental platform for drone state estimation under non-inertial reference frames. The setup includes dual motion capture systems, a mobile cart, and a Crazyflie 2.0 micro aerial vehicle integrated with ROS-based control.
 }
  \label{fig:setup}
\end{figure}

To evaluate the proposed control strategy under non-inertial conditions, we designed a series of experiments using a Crazyflie 2.0 nano quadrotor platform. The Crazyflie is equipped solely with its onboard IMU, and no additional sensors were used. Motion tracking was performed using two independent motion capture (MoCap) systems: a small-scale OptiTrack system (Flex 3, 100 Hz) mounted on a mobile cart, and a large-scale Qualisys system installed in the lab environment. The OptiTrack system provided non-inertial pose feedback of the drone relative to the cart, while the Qualisys system offered global inertial ground truth of the cart's motion as illustrated in Fig. ~\ref{fig:setup}. 
Figure~\ref{fig:setup} illustrates the architecture of the experimental platform. The system includes two motion capture systems (OptiTrack and Qualisys) mounted in different reference frames to simultaneously track the drone and the cart. Reflective markers are attached to both the Crazyflie 2.0 and the mobile platform. All pose data are transmitted via a local network and processed on a ROS-based computer, enabling synchronized control and logging. This hardware layout supports robust testing of state estimation algorithms in dynamic, non-inertial scenarios.

The mobile platform $(outer: x_c=1.8m, y_c=1.0m, z_c=1.0m)$, referred to as the \textit{cart}, served as a surrogate for a moving base (e.g., a vehicle). While exact dimensions are deferred, the drone flew within a bounded volume denoted as $(inner: x_f=1.7m, y_f=0.9m, z_f=0.9m)$. The control system and data logging were implemented in ROS, and all position, velocity, and estimator states were recorded during flight. For comparative evaluation, we implemented both our method and a baseline Extended Kalman Filter (EKF)-based estimator. 
To ensure proper temporal alignment between measurements, we employed a timestamp-based synchronization scheme. The drone's position and velocity were captured using an OptiTrack motion capture system via the VRPN ROS topic. Simultaneously, the platform (referred to as the cart) was tracked using a separate Qualisys motion capture system, publishing velocity and position under another ROS Qualisys topic. In the main pose callback, we aligned the cart data with drone measurements by checking the timestamp difference and only logging cart states when they were within a 100 ms window of the drone's motion capture timestamp. This lightweight synchronization strategy ensured consistency across heterogeneous sensor streams without requiring strict hardware clock alignment.

In this work, we depart from the traditional quadratic model which showed in Equation~\eqref{eq:u_combined}. Instead, we adopt a more accurate thrust model derived from empirical data provided by Bitcraze~\cite{bitcraze2025}, which characterizes the mapping between PWM duty cycle and measured thrust under different battery voltages. Using this dataset, we fit a smooth curve to obtain a continuous function for thrust as a function of PWM, and apply interpolation to compute the actual thrust generated by each rotor.




\subsection*{Experiment Types and Motion Modes}

To evaluate estimator robustness across different motion conditions, we designed three categories of experiments with increasing dynamic complexity. For all trials, the drone was commanded to maintain a fixed altitude of $0.35\,\mathrm{m}$. The large-scale Qualisys motion capture system was used to record cart states, ensuring high-fidelity ground truth for comparison. All test conditions used comparable platform acceleration magnitudes to facilitate fair evaluation across different motion modes.

\begin{itemize}
    \item \textbf{Experiment 1: Stationary Hover Test} — The cart remained completely still throughout the flight. This served as the baseline case for evaluating estimator behavior under ideal, disturbance-free conditions. Both estimators were expected to produce minimal velocity errors and stable state estimates, as no platform motion was present.

    \item \textbf{Experiment 2: $X$-Axis Forward Motion Test} — The cart moved forward along the $x$-axis under two speed profiles corresponding to moderate and high accelerations. While the platform motion was manually actuated, the velocity was recorded using the Qualisys system to ensure repeatability and consistency across trials. This setup tested the estimators’ ability to reject translational disturbances while tracking a moving platform.

    \item \textbf{Experiment 3: $X$ and $Y$-Axis Motion Test} — To evaluate performance under not only $x$-axis, the cart was yawed by $45^\circ$ after takeoff, with the drone maintaining its yaw aligned to the world $x$-axis (yaw control disabled). This misalignment caused the drone to fly diagonally in the cart frame, requiring the estimator to resolve motion in both $x$ and $y$ directions under a non-aligned reference frame. Two different platform speeds were tested under this condition, both having similar accelerations to those in Experiment 2.
\end{itemize}

\section{Results and Discussion}

We compare the performance of the proposed state estimation approach with the standard Extended Kalman Filter (EKF) under static conditions, using the same controller for both methods. The experimental routine involves take-off, hovering at a target altitude of 0.35 meters while the $x$ and $y$ axis is still zero for approximately 15 seconds, followed by an automated landing.

\begin{figure}[!t]
  \centering
  \includegraphics[width=1.0\linewidth]{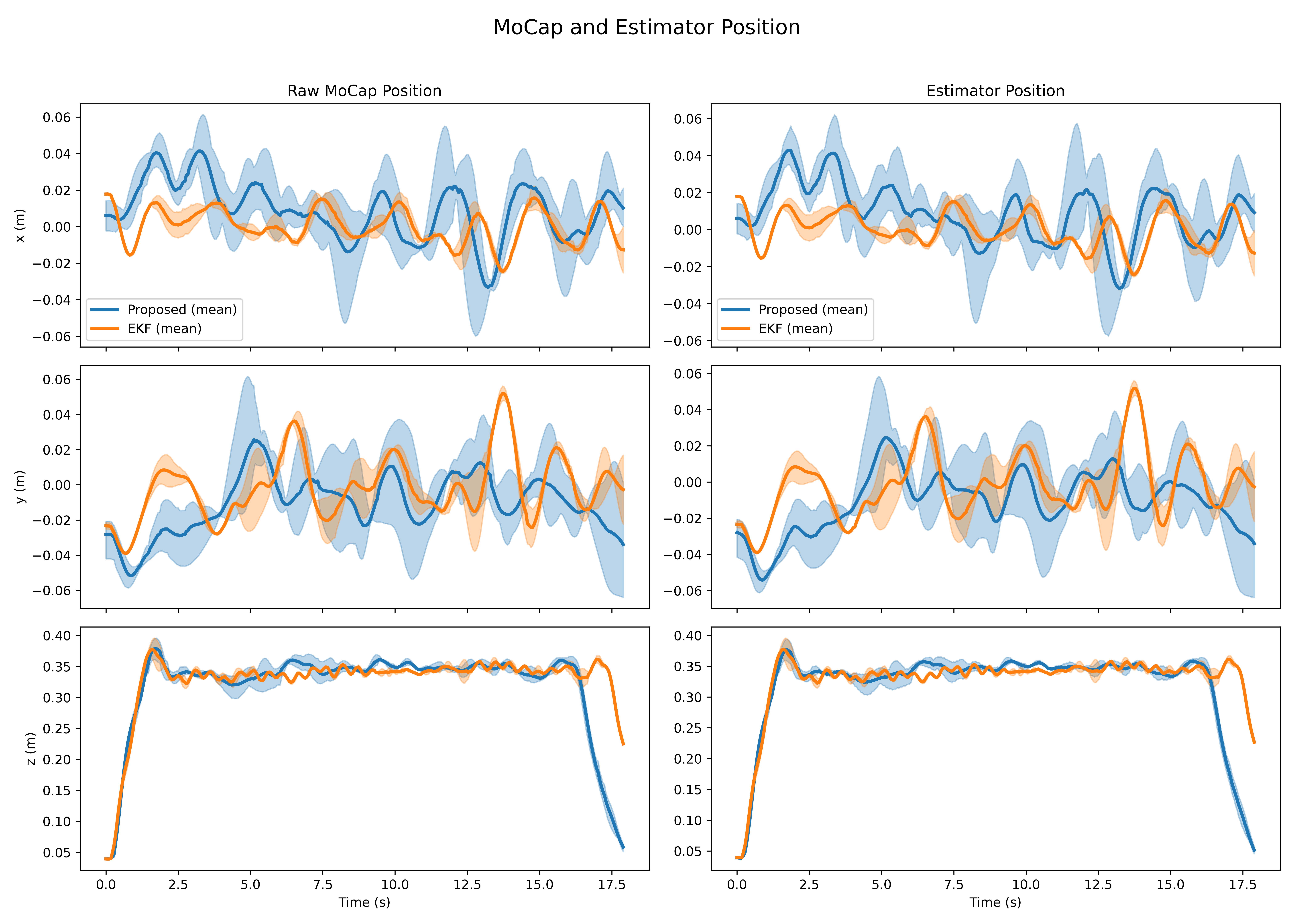}
  \caption{
    Experiment~1 (Stationary Hover Test): Comparison of 3D position estimation.
    The blue curves correspond to the proposed EKF-UI method, while the red curves denote the baseline standard EKF.
    The first, second, third row shows the position along the $x$-axis, $y$-axis, and the $z$-axis.
  }
  \label{mocapraw}
\end{figure}

\begin{figure}[!t]
  \centering
  \includegraphics[width=1.0\linewidth]{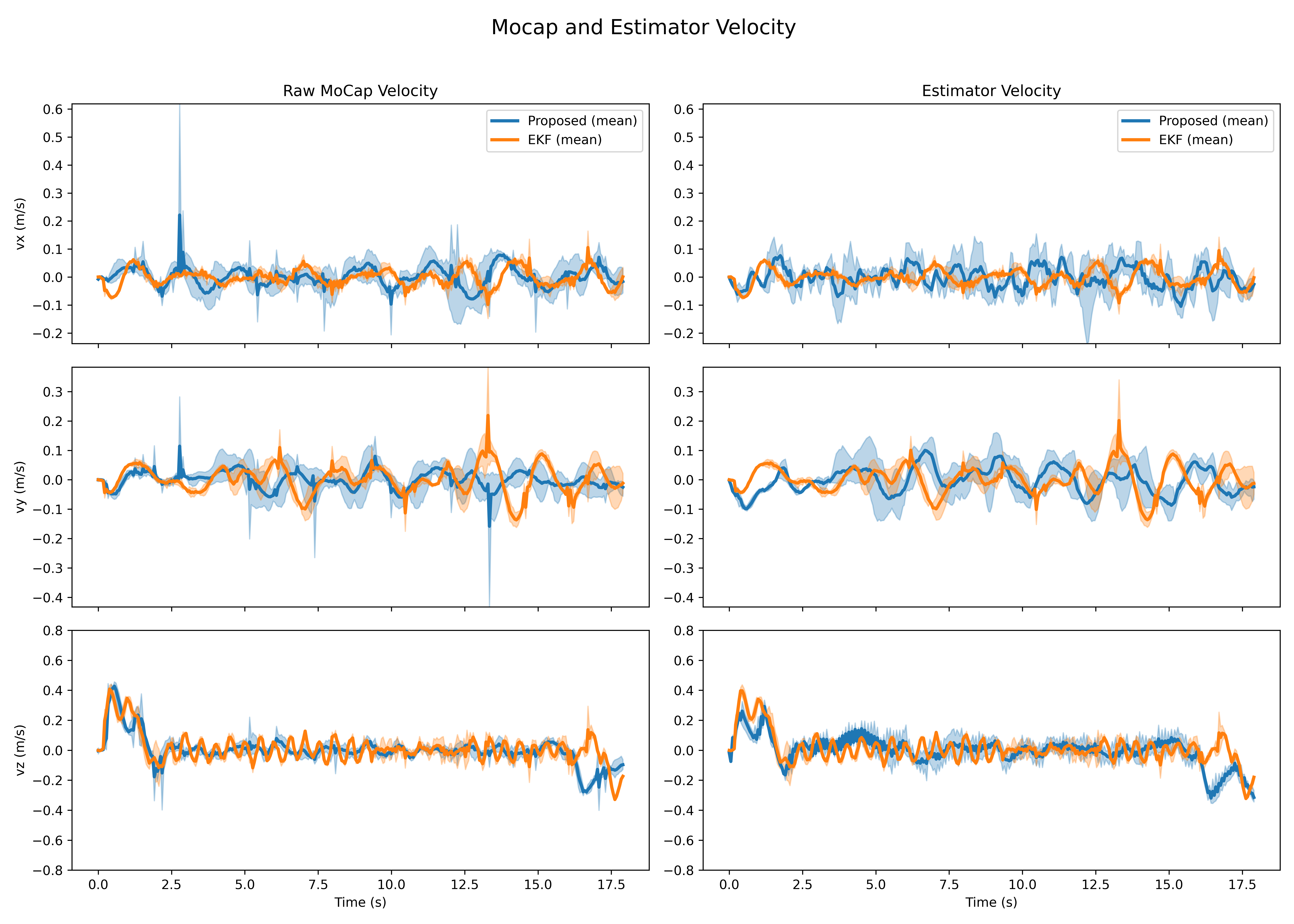}
  \caption{
    Experiment~1 (Stationary Hover Test): Comparison of 3D velocity estimation.
    The blue curves represent the proposed EKF-UI method, whereas the red curves correspond to the baseline standard EKF.
    The first, second, third row presents the velocity along the $x$-axis, $y$-axis, and the $z$-axis.
  }
  \label{mocaprawv}
\end{figure}

As illustrated in Fig.~\ref{mocapraw} and Fig.~\ref{mocaprawv}, we present a detailed comparison between the proposed state estimation method (blue) and the baseline EKF (orange). The left columns in both figures show raw motion capture (MoCap) data while the right columns depict the corresponding estimator outputs from each method.

Specifically, Fig.~\ref{mocapraw} contrasts the 3D position trajectories directly obtained from MoCap measurements (left) with those estimated onboard (right). Fig.~\ref{mocaprawv} shows the corresponding velocity data: on the left, velocities are derived by numerically differentiating MoCap positions, while on the right, the estimates come directly from each state estimator. In all subplots, both methods are shown with their average trajectories and shaded regions indicating trial-to-trial variability.

Notably, in Fig.~\ref{mocaprawv}, the raw differentiated velocity signals exhibit visible high-frequency noise and sharp spikes. The proposed method (blue) effectively smooths these fluctuations in its estimated outputs, particularly in the vertical velocity ($v_z$). This improvement stems from the incorporation of physical modeling and dynamic constraints in the estimator design, which suppresses measurement noise while maintaining fidelity to the actual motion.

Overall, both methods provide reliable state estimates during static hover. The proposed method demonstrates accuracy comparable to the EKF in both position and velocity, even when relying solely on raw MoCap input. These results validate the effectiveness of the proposed approach and highlight its potential for future use in more challenging, dynamic, and non-inertial conditions.

\begin{figure}[!t]
  \centering
  \includegraphics[width=1.0\linewidth]{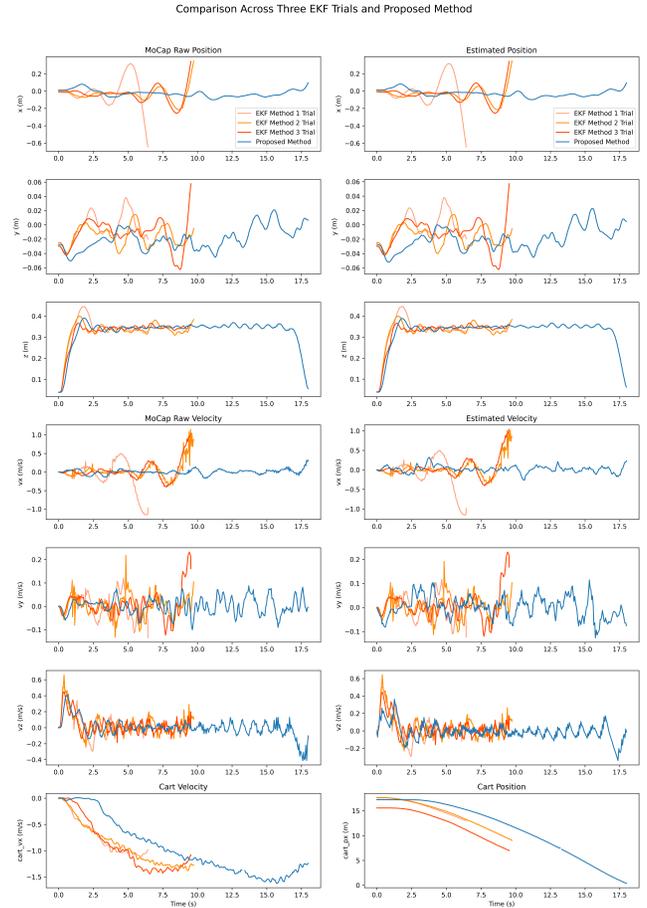}
  \caption{ Experiment~2 ($X$-Axis Forward Motion Test):
    Comparison of three EKF-based estimation trials (orange variants) and the proposed method (blue) for drone state estimation and cart motion tracking. 
    Each row presents a different category of state information: (Top three rows) MoCap raw and estimated positions in $x$, $y$, and $z$ directions; 
    (Middle three rows) corresponding velocities; 
    (Bottom row) cart’s longitudinal velocity and position.
    All plots share a synchronized time axis, and the proposed method demonstrates smoother estimation performance with reduced noise and tighter alignment across trials.
  }
  \label{fig:compare_3in1}
\end{figure}

In the Forward Motion Test:
As illustrated in Fig.~\ref{fig:compare_3in1}, the proposed method consistently outperforms the three EKF-based baseline trials in terms of estimation smoothness and robustness, especially in raw velocity and cart motion tracking.

\begin{figure}[!t]
  \centering
  \includegraphics[width=1.0\linewidth]{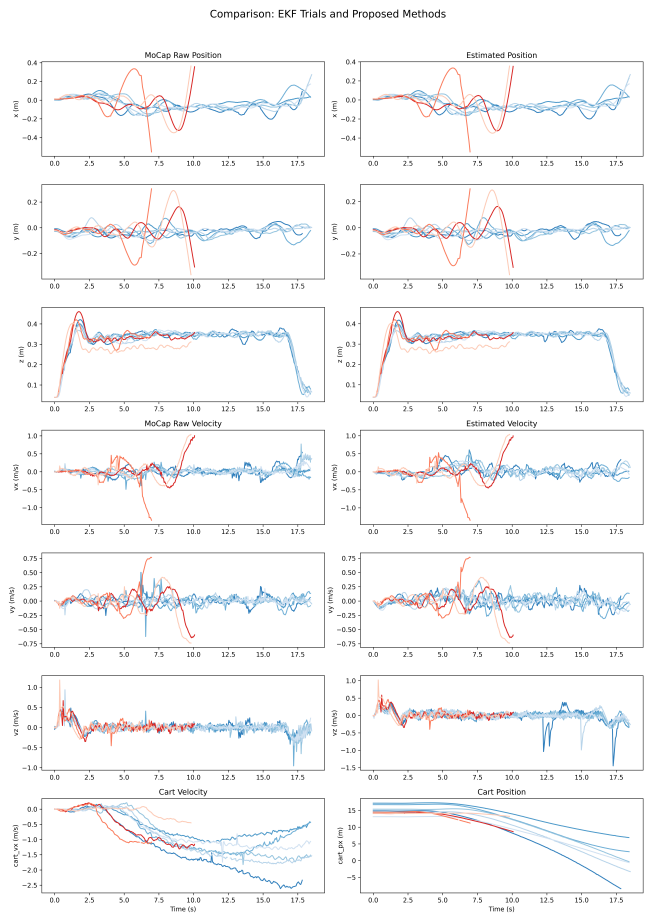}
  \caption{ Experiment~3 ($X$ and $Y$-Axis Motion Test):
    Comparison of three EKF-based estimation trials (red variants) and the proposed method (blue variants) under a yaw-induced dissonance scenario.
    Each row presents a different class of state information: 
    (Top three rows) MoCap raw and estimated positions in $x$, $y$, and $z$ directions;
    (Middle three rows) corresponding raw and estimated velocities; 
    (Bottom row) cart’s longitudinal velocity and position.
    All plots share a synchronized time axis. 
    Red lines represent the three EKF trials, while blue lines represent the proposed method across multiple trials.
    The EKF results are plotted on top for visual emphasis. 
  }
  \label{fig:compare_yaw_dissonance}
\end{figure}

To evaluate estimator performance under both $x$ and $y$ axis, we designed an experiment where the mobile cart was yawed by $45^\circ$ shortly after drone takeoff. Importantly, the drone's yaw control was disabled throughout the trial, keeping its body frame aligned with the global $x$-axis. As the cart rotated beneath the drone, this misalignment caused the drone to fly diagonally relative to the cart frame, effectively decoupling translational motion in the body and world frames.

This setup served as a critical test of the estimators' ability to track motion in a non-inertial, misaligned reference frame. Different cart speeds were tested to further explore estimator robustness under varied dynamic conditions our proposed method resulting in consistent tracking, and the stander EKF revealing limitations of the baseline EKF approach.

Figure~\ref{fig:compare_yaw_dissonance} shows the comparative results from three EKF trials and multiple runs using our proposed method. As seen in the plots, the proposed method consistently delivers smoother estimates with reduced jitter in both position and velocity, particularly under the $x$-$y$ motion. Additionally, the cart tracking performance remains tight across trials, further confirming the robustness of the proposed estimator in both $x$ and $y$ axis.

\begin{figure}[!t]
  \centering
  \includegraphics[width=1.0\linewidth]{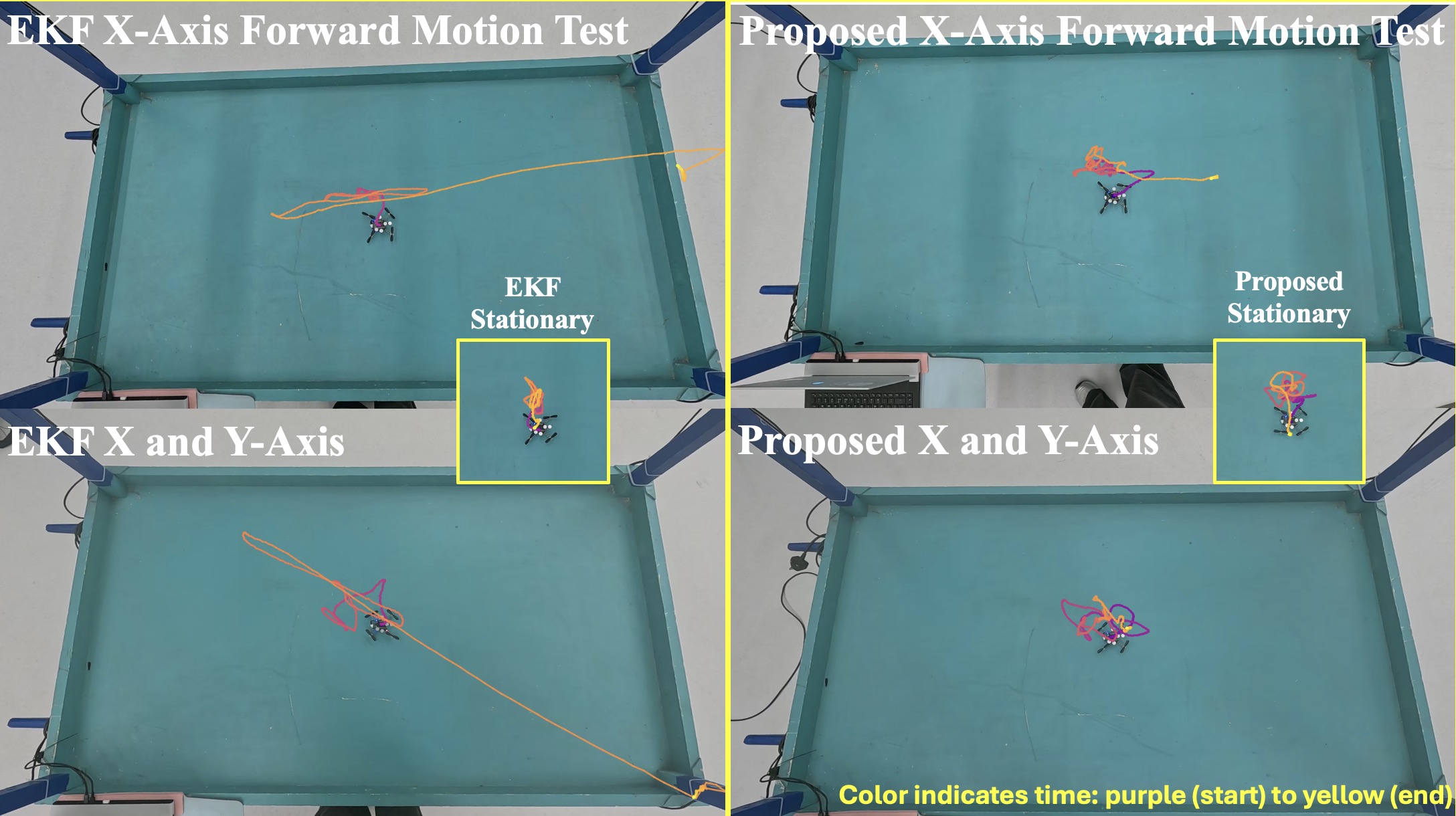}
  \caption{
    Comparison of drone trajectories under both static and dynamic conditions using the baseline EKF (left column) and the proposed method (right column). 
    The layout is organized as follows: the two central insets correspond to Experiment~1 (Stationary Hover Test), the top row shows Experiment~2 (Standard Movement Test), and the bottom row depicts Experiment~3 (Yaw-Disoriented Test), where the cart was rotated by $45^\circ$ after takeoff without yaw control on the drone. 
    In both dynamic experiments (Experiments~2 and~3), the proposed method yields tighter and more consistent trajectory tracking compared to the baseline EKF, with the most pronounced improvements observed during diagonal motion in the $45^\circ$ condition. 
    The static hover tests (insets) demonstrate that both methods achieve comparable performance when the platform remains stationary. 
    Color indicates time progression from purple (start) to yellow (end).
  }
  \label{fig:traj_compare_full}
\end{figure}

Figure~\ref{fig:traj_compare_full} compares drone trajectories in both $X$-Axis Forward Motion Test and $X$ and $Y$-Axis Motion Test(45° yaw rotated trials). The proposed method demonstrates significantly better tracking in both cases, especially under diagonal motion, while matching EKF performance during static hovering.

\section{Conclusion}

This paper presents a robust control and estimation strategy for quadrotors operating in GPS-denied, non-inertial environments, where conventional inertial sensing becomes unreliable due to unknown platform dynamics. To address the challenge of velocity estimation in non-inertial frames, we integrate an Extended Kalman Filter with Unknown Inputs (EKF-UI) that estimates platform-induced accelerations online and provides cleaner velocity feedback to the controller.

Our experimental platform, built around the Crazyflie 2.0 nano drone, employs dual motion capture systems (OptiTrack and Qualisys) to independently track the drone and the moving platform. Across three distinct experimental conditions such as static hover, $x$ axis forward translation, and $x$ and $y$ motion, we validate the control system’s robustness under increasing dynamic complexity. The proposed controller consistently outperforms conventional EKF-based systems, demonstrating smoother responses, reduced tracking errors, and enhanced stability.

These findings highlight that in dynamic, non-inertial scenarios, the control architecture must be designed with awareness of external platform disturbances. Rather than treating estimation and control as separate modules, our approach tightly integrates a disturbance-aware estimator to serve the needs of the controller. This framework offers a low-complexity solution for deploying UAVs on moving platforms, and provides a practical pathway for robust estimation-based control in challenging environments.

\begin{acks}
This class file was developed by Sunrise Setting Ltd,
Brixham, Devon, UK.\\
Website: \url{http://www.sunrise-setting.co.uk}
\end{acks}

\bibliographystyle{TRR}
\bibliography{refs}

\end{document}